\documentclass[sigconf]{acmart}
\renewcommand\footnotetextcopyrightpermission[1]{} 
\usepackage{booktabs} 

\usepackage[T1]{fontenc}  

\usepackage{url}            

\usepackage{breakurl}
\usepackage[breaklinks]{hyperref}

\usepackage{booktabs}    
\usepackage{amsfonts}    
\usepackage{nicefrac}    
\usepackage{microtype}   

\usepackage{graphicx} 
\graphicspath{{./images/}}
\DeclareGraphicsExtensions{.pdf,.jpeg,.png}

\usepackage{subfig}
\usepackage{algorithm}
\usepackage{algorithmic}

\usepackage{subfiles}
\usepackage[subpreambles=false]{standalone}

\usepackage{bm}
\usepackage{subfiles}
\usepackage{wrapfig}
\usepackage{epsfig}
\usepackage{caption}

\graphicspath{{./images/}}
\DeclareGraphicsExtensions{.pdf,.jpeg,.jpg,.eps,.png}

\newcommand{\INST}[1]{}

\newcommand{\KL}[1]{\textbf{KL}}

\newcommand{\FvF}{\emph{Fly-vs-Fly}}
\newcommand{\BB}{\emph{BBShot}}
\newcommand{\BBmini}{\BB{}-200k}

\usepackage{amsmath, amsthm, amssymb, xspace}

\usepackage{color}
\definecolor{blue}{rgb}{0,0,.7}
\definecolor{red}{rgb}{.7,0,0}
\definecolor{orange}{rgb}{1,.6,0}
\definecolor{purple}{rgb}{.4,0,.5}
\definecolor{brown}{rgb}{.4,.2,.1}
\definecolor{green}{rgb}{0,.5,0}

\newcommand{\refn}[1]{(\ref{#1})}

\newcommand{\brck}[1]{\left(#1\right)}
\newcommand{\brcksq}[1]{\left[#1\right]}
\newcommand{\brckcur}[1]{\left\{#1\right\}}

\newcommand{\brcka}[1]{\langle #1\rangle}

\newcommand{\fr}[2]{\frac{#1}{#2}}

\newcommand{\be}{\begin{equation}}
\newcommand{\ee}{\end{equation}}
\newcommand{\bali}{\begin{eqnarray*}}
\newcommand{\eali}{\end{eqnarray*}}
\newcommand{\eq}[1]{\begin{align}#1\end{align}}
\newcommand{\eqn}[1]{\begin{align*}#1\end{align*}}

\newcommand{\iitem}[1]{\begin{itemize}#1\end{itemize}}

\newcommand{\calA}{\mathcal{A}}

\newcommand{\calO}{\mathcal{O}}

\newcommand{\mathR}{\mathbb{R}}

\newcommand{\mathE}{\mathbb{E}}

\newcommand{\tb}[1]{\textbf{#1}}
\newcommand{\ti}[1]{\textit{#1}}

\newcommand{\argmi}[1]{\textrm{argmin}_{#1}}

\newcommand{\ddd}[1]{^{{#1}}}
\newcommand{\xx}{\mathbf{x}}

\newcommand{\ww}{\mathbf{w}}

\newcommand{\algMMT}{\texttt{MRTL}}
\newcommand{\algSGD}{\texttt{SGD-se}}
\newcommand{\algFinegrain}{\texttt{Finegrain}}
\newcommand{\algTensorFactorize}{\texttt{Factorize}}

 \usepackage{mathtools}
 \usepackage{color}

\newcommand{\loss} {\mathcal{L}}

\newcommand{\eat}[1]{}

\newtheorem{theorem}{Theorem}[section]

\newtheorem{lemma}[theorem]{Lemma}

\newcommand{\R}{{\mathbb R}}
\newcommand{\E}{{\mathbb E}}

\newcommand{\T}[1]{{\mathcal{#1}}} 
\newcommand{\V}[1]{{\mathbf{#1}}} 

\setcopyright{none}

\begin{document}
\title{Multi-resolution Tensor Learning for Large-Scale Spatial Data }

\author{Stephan Zheng}
\affiliation{%
 \institution{California Institute of Technology}
 \streetaddress{1200 E California Boulevard}
 \city{Pasadena}
 \state{California}
 \postcode{91125}
}
\email{stephan@caltech.edu}
\author{Rose Yu}
\affiliation{%
 \institution{California Institute of Technology}
 \streetaddress{1200 E California Boulevard}
 \city{Pasadena}
 \state{California}
 \postcode{91125}
}
\email{rose@caltech.edu}
\author{Yisong Yue}
\affiliation{%
 \institution{California Institute of Technology}
 \streetaddress{1200 E California Boulevard}
 \city{Pasadena}
 \state{California}
 \postcode{91125}
}
\email{yyue@caltech.edu}

\renewcommand{\shortauthors}{S. Zheng et al.}

\begin{abstract}
High-dimensional tensor models are notoriously computationally expensive to train.
We present a meta-learning algorithm, \algMMT{}, that can significantly speed up the process for spatial tensor models.
\algMMT{} leverages the property that spatial data can be viewed at multiple resolutions, which are related by coarsening and finegraining from one resolution to another.
Using this property, \algMMT{} learns a tensor model by starting from a coarse resolution and iteratively increasing the model complexity.
In order to not "over-train" on coarse resolution models, we investigate an information-theoretic fine-graining criterion to decide when to transition into higher-resolution models.
We provide both theoretical and empirical evidence for the advantages of this approach.
When applied to two real-world large-scale spatial datasets for basketball player and animal behavior modeling, our approach demonstrate 3 key benefits:
1) it efficiently captures higher-order interactions (i.e., tensor latent factors),
2) it is orders of magnitude faster than fixed resolution learning and scales to very fine-grained spatial resolutions, and
3) it reliably yields accurate and interpretable models.
%
%


\end{abstract}

%
%



\maketitle

\begin{acks}
This result is supported in part by NSF \#1564330, NSF \#1637598, and gifts from Bloomberg and Northrop Grumman.
\end{acks}

\section{Introduction}
\label{sec:intro}

We study the problem of learning high-dimensional tensor models from large-scale high-resolution spatial data.
Such models can compactly describe \emph{multi-way} correlations between predictive features that are spatially distributed.
%
%
For example, in competitive basketball play, given the positions of all players in the court, we can predict \emph{whether a player will shoot at the basket} using a high-dimensional tensor model.
Previous research on individual player models has demonstrated the effectiveness of matrix latent factor models \cite{Yue2014}.
For joint team behavior, we can model a basketball player's profile by learning the latent factors among offensive team positions,
defending team positions and the player's own profile, which can be modeled by a \emph{tensor latent factor model}.


This paper addresses two major challenges for learning spatial tensor models.
First, we study \emph{scalability}, as high resolution spatial tracking data is large-scale.
Computational methods usually involve fine-grained spatial discretization, leading to high-dimensional tensors that are computationally slow and expensive to train.

In addition to scalability, we are also interested in learning \emph{interpretable} models.
Apart from accurate prediction, we also wish to learn latent factors that can provide insights for domain experts.
For instance, the latent factors in basketball player models could correspond to player performance profiles, revealing hidden traits and commonalities among players.

\begin{figure}[!t]
\centering
\includegraphics[width=0.45\linewidth]{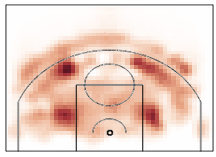}
\includegraphics[width=0.45\linewidth]{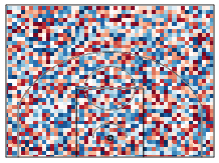}
\caption{
Comparison of learned latent factors for basketball shot prediction.
Left: interpretable latent factors learned by our multi-resolution method.
Right: uninterpretable factors learned by learning at fixed resolution.
Despite similar accuracy, the uninterpretable profile is noisy and not spatially smooth, whereas the interpretable profile shows basketball shooting hotspots that are relevant for prediction.}
\label{fig:bb_bad_init_noise}
\end{figure}

Towards scalable learning of interpretable tensor latent factor models, a key technical challenge is their non-convex nature \cite{yu2016learning}: optimization algorithms can converge to different local optima, depending on the model initialization.
Hence, a good initialization is critical for obtaining accurate and interpretable latent factors.
For instance, Figure \ref{fig:bb_bad_init_noise} shows two basketball shooting profiles with similar prediction accuracy, learned using the same optimization algorithm (gradient descent).
Using a good initialization yields an interpretable solution, while a random initialization does not.

\paragraph{Our Contributions.}
Thus we are interested in answering how we can  \ti{efficiently} learn both \ti{accurate} and \ti{interpretable}  tensor models from high-resolution spatial data.
In this paper, we present a novel meta-learning algorithm: multi-resolution tensor learning (\algMMT{}), that can 1) learn a good initialization, 2) automatically control when to fine-grain, and 3) easily scale to high-dimensional tensors.
More importantly, the multi-resolution learning scheme yields interpretable solutions that are spatially smooth and relevant for the prediction tasks.

\algMMT{} is based on three key insights.
First, to obtain good initializations, instead of directly learning the latent factors from a random initialization, we can first learn a full-rank parameter tensor and use the factors from this parameter tensor as initialization. An analogous approach for matrix latent factor models has been shown in \cite{Miller2014,Yue2014} to yield interpretable factors capturing cohesive spatial semantics.

Second, to avoid the curse of dimensionality, we leverage the characteristics of spatial data and learn iteratively at \ti{multiple resolutions}.
At a high-level, the multi-resolution training has two stages: in the first stage, we learn a full tensor model using multi-resolution and factorize it as initialization for the latent factor model.
In the second stage, we train the latent factor model using multi-resolutions: starting at a coarse resolution and fine-graining the latent factors during training.
We prove that the resulting algorithm is faster by a logarithmic factor of the Lipschitz constant of the objective function and the estimation error.

Third, we investigate fine-graining criteria.
One simple criterion is to measure training loss convergence.
However, it requires a preset threshold, which does not easily reflect the spatial distribution at different granularities.
We thus take an information theoretical approach by monitoring the entropy of the gradients distribution for every grid cell, which we refer to as \ti{spatial entropy}.
We found this fine-graining criterion to be more effective than loss convergence.


In summary, our main contributions are:
\iitem{
\item We present \algMMT{}: a multi-resolution meta-algorithm to learn spatial latent factor models and a number of instantiations using various fine-graining criteria.
\item We theoretically analyze \algMMT{} and show that it converges faster to a similar accuracy than training at a fixed resolution.
\item We empirically show that using gradient statistics, (e.g. gradient entropy), is an effective transition control method across hyperparameters and can outperform loss convergence.
\item We empirically demonstrate orders-of-magnitude faster learning than conventional fixed-resolution training.
\item We show that our approach reliably and efficiently yields interpretable spatial latent factor models on real-world basketball and animal tracking data.
}

\section{Related Work}
Modeling spatial data enjoys a long history in the data mining community \cite{miller2009geographic}. Recent applications include urban air quality prediction \cite{hsieh2015inferring}, social media recommendation \cite{yin2016spatio}, and event detection \cite{zhao2016hierarchical}, among others. In this work, we study multi-agent tracking data extracted from high-fidelity tracking cameras. Such spatial data contains high resolution information and are large-scale (cameras operate at high frequency and every frame is one data point).

Latent factor models have been a popular method for spatial data modeling \cite{reich2010latent,Yue2014,deng2016latent}. By projecting the raw data into a low-dimensional space, latent factor models can encode patterns such as spatial clustering. However, most existing latent factor models have only focused on capturing two-way interactions (such as user-location matrix models \cite{koren2009matrix}). For multi-agent behavior modeling, we extend the matrix latent factor model to capture multi-way correlations among agents, resulting in a tensor latent factor model. For instance, we can build a ``player $\times$ offense position $\times$ defense position'' tensor model for basketball plays.

Tensor models have recently gained considerable attention (cf. \cite{Jenatton2012,takeuchi2013non,Quattoni2014}), but are still of limited applicability due to computational, data sparsity, and memory efficiency concerns. There are many efforts to scale up tensor computation, such as random projection (sketching) \cite{wang2015fast}, parallel computing \cite{austin2016parallel}, or utilizing sparsity structure \cite{perros2017spartan}. In this work, we take a top-down approach to learn a tensor model at multiple resolutions. The multi-resolution training procedure takes advantage of the spatial characteristics of the raw data, and is generically applicable to different tensor models and optimization algorithms.

Our approach bears affinity to other multi-stage meta-algorithms (e.g., \cite{johnsonblitz}) and the multi-grid method in PDE analysis. For instance, \citep{chow1991optimal} studies multi-grid methods in the context of dynamic programming. In the machine learning community, our method is also related to multi-resolution sparse approximation \cite{mallat1989multiresolution}. For example, \cite{kondor2014multiresolution} studies multi-resolution matrix factorization based on wavelets theory. \cite{schifanella2014multiresolution} proposes a multi-resolution heuristic for tensor factorization. In contrast, we study multi-resolution learning for \emph{tensor} models and also provide theoretical analysis for its convergence.

\section{Tensor Latent Factor Models}
%
We motivate the tensor latent factor model using the example of competitive basketball play. Suppose we have $n_a$ players and want to predict whether a basketball player $a$ ($a=1\ldots n_a$) takes a shot at the basket (binary label $y_a = \pm 1$).
We first discretize the court using a grid with $m$ cells (positions $\V{x} \in \R^m$) and then construct a binary feature vector $\phi(\xx)\in\mathbb{R}^m$ using the positions of all other players on the court.
The prediction problem can be formulated as:
\eq{\label{eq:linear}
P(y_a = 1 | \xx) = \brcka{\ww_a,\phi(\xx)} +\V{b}_a, \hspace{10pt}
\phi(\xx) = (0 \ldots 1 \ldots 0),
}
where $\ww_a \in \R^m$ are the weight parameters for the grid cells and $\V{b}_a$ is the bias unit. The weights $\ww_a$ encodes how likely a basketball player will shoot from a given position on the basketball court.

\begin{figure}[t]
	\centering
	\includegraphics[width=0.8\linewidth]{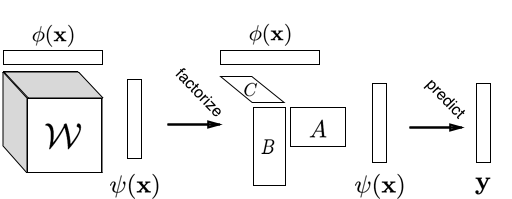}
	\vspace{-0.05in}
	\caption{Prediction process using tensor latent factor model, from full-rank tensor to latent factor model.}
	\label{fig:model}
	\vspace{-0.13in}
\end{figure}

\paragraph{Matrix Latent Factor Models.}
In matrix form, (\ref{eq:linear}) is equivalent to the following model:
\eq{\label{eq:matrix}
	f_a(\xx) 
	&= \sum_{b} W_{ab} \phi_{b}(\xx)+ \V{b}_a,
}
where $f_a(\xx)$ is the prediction function for player $a$, and the matrix $W$ concatenates the parameter vectors $\{ \V{w}_a \}$. Latent factor models assume there exist low-dimensional representations of the weights. Hence the parameter matrix factorizes into two matrices $A$ and $B$ with $K$ components:
\eq{W_{ab} = \sum_k A_{ak} B_{bk}.}
These latent factors can represent players' behavioral profiles at different positions on the court.
\paragraph{Tensor Latent Factor Models.}
The limitation of the matrix latent factor model in (\ref{eq:matrix}) is that it can only capture the correlation between the ballhandler and all other players.
In order to learn rich behaviors between teams, we need to explicitly model multi-way dependencies and generalize to high-order models.
In the basketball example, we separately construct a feature vector $\phi(\xx), \psi(\xx) \in\mathbb{R}^d$ for the offense and defense team positions, respectively, and model the multi-way interactions as:
\eq{\label{eq:tensor}
	f_a(\xx) 
	&= \sum_{bc} \T{W}_{abc} \phi_{b}(\xx) \psi_{c}(\xx) + \V{b}_a.
}
%
Here $\T{W}$ is an order 3 weight tensor.
To encode low-dimensional structure, we assume the weight tensor $\T{W}$ factorizes:
\eq{\T{W}_{abc} = \sum_{k=1}^K A_{ak}B_{bk}C_{ck},}
which corresponds to a CP tensor model \cite{Kolda2009}. In this case, $K$ is called the \ti{tensor rank}.
In general, given some input $\xx$ and feature transformation functions $\phi(\cdot)$ and $\psi(\cdot)$,
\ti{tensor latent factor models} aim to learn a function $f_a(\xx)$ for each prediction task $a$, such that:
\eq{\label{eq:tucker}
f_a(\xx) 
&= \sum_{bc} \sum_{k=1}^K A_{ak}B_{bk}C_{ck} \phi_{b}(\xx)\psi_{c}(\xx) + \V{b}_a,
}
where $A, B, C$ are the latent factors, $\V{b}_a$ is a bias unit and $a$ indexes prediction tasks. It is straightforward to generalize (\ref{eq:tucker}) to other tensor models and higher orders. Figure \ref{fig:model} depicts the model details.

The dimension of the weight tensor relates to the discretization size of the spatial data. The main motivation for considering discretizations is that they allow us to learn flexible non-parametric models.
In contrast, traditional parametric models such as spatial point processes \cite{Miller2014} have strong assumptions on the form of the spatial correlations (e.g. the generating process is a distribution $P(y|x)$ with multiple modes or has a cyclic factorization structure), which can be hard to learn using conventional learning methods.

Tensor models explicitly capture higher-order correlations among the tasks and spatial features. The learned latent factors can also be more interpretable due to the multi-linear nature (in the spatial feature dimensions).
For example, if we see the prediction function $f(\xx) = P(y|\xx)$ as a distribution and $\phi_b(\xx), \psi_c(\xx)$ are 1-hot vectors that encode spatial occupancy, the columns $B_{\cdot k}$ and $C_{\cdot k}$ (for a fixed $k$) can be interpreted as smooth spatial probability distributions, representing players' shooting profile.
%

The cost of using such models, however, is that they lead to a hard non-convex optimization problem and can suffer from multiple local optima. Moreover, different initializations can lead to more or less interpretable models.
They are also computationally expensive to train, making them infeasible for large-scale high resolution spatial data.
Hence, the goal of our multi-resolution method is to efficiently find accurate and interpretable solutions in this setting.




%
\section{Multi-resolution Learning}
\label{sec:mrl}
We now describe our approach for training interpretable spatial latent factor models, such as the tensor model $\T{W}$ in \refn{eq:tensor}. In this section, for simplicity we consider only fine-graining a single dimension of $\T{W}$, although our analysis can be easily generalized.

\paragraph{Initialization from Factorization.} The non-convex nature of the tensor model requires a good initialization. We obtain this by factorizing a trained full-rank tensor model as in \refn{eq:tucker} and using the factors as initializations. This approach has been shown to be effective in learning predictive spatial patterns \cite{Miller2014,Yue2014}.  We also have similar observations in our experiments.


\paragraph{Iterative Fine-graining.}
It is generally not tractable in memory to learn a high-dimensional tensor latent factor model, that operates at some high resolution. To make training feasible, the main idea of our algorithm is iterative fine-graining.

We will denote the \emph{resolution (i.e. size of a grid cell)} as $d_i$, which is inversely proportional to the number of grid cells $m_i$.

During the first (full-rank) phase, we start training with a \ti{full-rank model} $\T{W}_{d_0}$ that operates at a coarse resolution $d_0$ and increase the spatial resolution of the data and model using a sequence of resolutions $d_0, d_1,\ldots,d_f$.
That is, instead of learning $\T{W} \in \R^{\ldots\times m_f \times \ldots}$, we learn a sequence of $\{\T{W}_{d_i} \in \R^{\ldots\times m_i \times \ldots} \}$.
When moving to a higher resolution, we use the learned weights $\T{W}_{d_i}$ to initialize the next weights $\T{W}_{d_{i+1}}$, by scaling up $\T{W}_{d_i}$ along the spatial dimension.

We leverage the fact that spatial data can be viewed at multiple resolutions. Lower resolution models can provide good initializations for higher resolution models with high accuracy.
To determine when to transition, we study different fine-graining criterion, which we discuss in section \ref{ss:sgd}.

\begin{figure}[t]
	\centering
	\includegraphics[width=\linewidth]{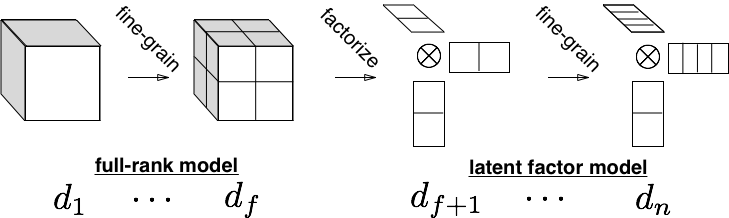}
	\vspace{-0.05in}
	\caption{Depicting our multi-stage training, which starts with a coarse-grained full-rank model, and concludes on a fine-grained latent factor model. See Algorithm \ref{alg:mmt} and Algorithm \ref{alg:sgd-se} for more details.}
	\label{fig:diagram}
	\vspace{-0.13in}
\end{figure}

In the second phase, we apply a tensor factorization to $\T{W}_{d_i}$ at resolution $d_f$ as initialization. We use a similar fine-graining procedure for the latent factor model to the final resolution $d_n$.
The resolution $d_f$ is determined by considering the memory requirements of the model at different resolutions.
This meta-algorithm is outlined in Algorithm \ref{alg:mmt} and depicted in Figure \ref{fig:diagram}.
\subsection{Computational Complexity Analysis}
We analyze the computational complexity for \algMMT{} (Algorithm \ref{alg:mmt}). Intuitively, as most of the training iterations are spent on coarser resolutions with fewer number of parameters, multi-resolution training is more efficient than fixed-resolution training.

In this section, we formalize this intuition and give a rigorous analysis of \algMMT{}. The analysis is based on the multi-grid method \cite{stuben2001review} commonly used in partial differential equation analysis.

We denote the label $y \in \R^{n_a}$, features $\phi(x) \in \R^{m}$, $\psi(x) \in \R^{m'}$ and weight tensor $\T{W} \in \R^{n_a\times m \times m'}$. The prediction model is:
\eq{
f(x;\T{W})_a = \sum_{ab} \T{W}_{abc}\phi(\xx)_b \psi(\xx)_c+b_a.
}
Denote the objective function as $\loss(x,y, f(x;\T{W}))$ ,
\algMMT{} (Algorithm \ref{alg:mmt}) aims to solve the optimization problem:
\begin{eqnarray}
\min_{\T{W}} \E _{(x,y)\sim P} [\loss(x,y, f(x;\T{W}))].
\end{eqnarray}
where $\loss$ is the logistic loss. \algMMT{} follows gradient descent:
\eq{
\T{W} \leftarrow \T{W}- \lambda \nabla \loss \T{W}.
}
We first start with the coarsest resolution $d_0$ (discretization size), compute an initial estimate $\T{W}_{d_0}$, and keep iterating until $\T{W}_{d_0}$  satisfies certain fine-graining criteria.
Then we replace $d_0$ with $d_0/2$ and use $\T{W}_{d_0}$ as an initialization for the next level.

In general, this iterative fine-graining procedure yields a sequence of discretization sizes $[d_0, d_1, \cdots, d_n]$. Suppose for each resolution $d$, we use the following as the fine-graining criterion:
\eq{
\| \T{W}^{t(d)} - \T{W}^{t(d)-1}\|  \leq \fr{C_0 d}{\alpha (1-\alpha)}.
\label{eqn:fine-grain}
}
where $t(d)$ is the number of iterations needed at level $d$. The algorithm terminates when the estimation error reaches $\fr{C_0 d}{(1-\alpha)^2}$. We now show that \algMMT{} reduces the number of computation steps.

This is done by first formulating a gradient descent update as a fixed point iteration operator $F$\footnote{\emph{Stochastic} gradient descent converges to a noise ball instead of a fixed point.}:
\eq{
\T{W} \leftarrow F( \T{W}), \hspace{10pt} F:=I-\lambda \nabla \loss.
}
Assume the objective function is Lipschitz continuous, with a contraction constant of $\alpha \in (0,1)$, meaning:
\eq{
\| F(\T{W} ) - F(\T{W}') \|\leq \alpha \| \T{W} - \T{W}'\|.
}
\begin{algorithm}[t!]
\caption{\algMMT{}: Memory-efficient multi-resolution training with spatial entropy control}
\label{alg:mmt}
\begin{small}
\begin{algorithmic}[1]
\STATE Input: Tensor weights $\T{W}_{d_0}$ (e.g. Equation \refn{eq:tensor}), data $D$, features $\Psi$.
\FOR{each resolution $d_i \in \brckcur{d_1, \ldots, d_f}$}
\STATE \# For definition of \emph{resolution}, see Section \ref{sec:mrl}.
\STATE \algSGD{}$\brck{\T{W}_{d_i}}$ \# see Algorithm \ref{alg:sgdse}.
\ENDFOR
\STATE \tb{TensorFactors}$_{d_f}$ = \algTensorFactorize($\T{W}_{d_f}$) \# e.g. $A,B,C$ in Equation \refn{eq:tucker}.
\FOR{each resolution $d_i \in \brckcur{d_{f+1}, \ldots, d_n}$}
\STATE \algSGD{}$\brck{\tb{TensorFactors}_{d_i}}$
\ENDFOR \\
\RETURN $\tb{TensorFactors}_{d_n}$
\end{algorithmic}
\end{small}
\end{algorithm}
\begin{algorithm}[t!]
\caption{\algSGD{}: Stochastic gradient descent with spatial entropy control}
\label{alg:sgd-se}
\begin{small}
\begin{algorithmic}[1]
\STATE Input: $\T{W}_{d}$, data-set $D$, features $\Psi$, length-$T$ rolling window gradient buffer $H$.
\WHILE{termination condition (e.g. Condition \refn{eq:ent_cond}) not true}
  \STATE Gradient descent step on $\T{W}_{d_i}$ using minibatch $B$
  \STATE Add gradient $\brckcur{g(x,y)}_{(x,y)\in B}$ to $H$
\ENDWHILE
\STATE \algFinegrain($\T{W}_{d_i}$) \\
\RETURN $\T{W}_{d_{i+1}}$
\end{algorithmic}
\end{small}
\label{alg:sgdse}
\end{algorithm}
The following main theorem characterizes the speed-up gained by multi-resolution training w.r.t. the contraction factor $\alpha$ and the terminal estimation error $\epsilon$.
\begin{theorem}
Suppose the fixed point iteration operator (gradient descent) for the optimization algorithm has a contraction factor (Lipschitz constant) of $\alpha$, the multi-resolution training procedure is faster than that of the fixed resolution algorithm by a factor of  $ \log\fr{1}{(1-\alpha) \epsilon}$, with $\epsilon$ as the terminal estimation error.
\label{thm:mmt}
\end{theorem}
We prove several useful Lemmas before proving the main Theorem \ref{thm:mmt}.
We first analyze the computational cost of the \emph{fixed-resolution} algorithm.
%
\begin{lemma}
Given a fixed point iteration operator with a contraction factor of $\alpha$, the computational complexity of a fixed-resolution training for a $p$-order tensor with rank $r$ is
\eq{
T = \calO\brck{
\fr{1}{|\log \alpha|} \cdot \log  \fr{1 }{(1-\alpha)\epsilon}
\cdot\brck{\fr{rp}{(1-\alpha)^2\epsilon}}}.
}
\label{lemma:fixed}
\end{lemma}
\proof At a high level, we prove this by choosing a small enough resolution $d$ such that the approximation error is bounded with a fixed number of iterations.
Let $\T{W}_d^\star$ be the optimal estimate at level $d$ and $\T{W}^t$ be the estimate at step $t$. Then we have
\eq{
\| \T{W}^\star -\T{W}^t \| \leq \| \T{W}^\star - \T{W}_d^\star \|  + \|\T{W}_d^\star - \T{W}^t  \| \leq \epsilon.
}
Choose a fixed resolution $d$ that is small enough such that
\eq{\| \T{W}^\star-\T{W}^\star_d\| \leq \fr{\epsilon}{2},}
then the termination criteria
%
$\| \T{W}^\star-\T{W}^\star_d\| \leq  \fr{C_0 d}{(1-\alpha)^2}$
%
gives
%
$d = \Omega ((1-\alpha)^2 \epsilon)$.
%
Initialize $\T{W}^0 =0$ and iterate over $t$ times such that:
\eq{
\fr{\alpha^t}{2(1-\alpha) } \|F(\T{W}^0) \| \leq \fr{\epsilon}{2},
}
Since $\T{W}^0 =0$, $\|F(\T{W}^0) \| \leq 2C$, we obtain that
\eq{
t\leq  \fr{1}{|\log \alpha|} \cdot \log\fr{2C}{(1-\alpha)\epsilon},
}
Note that for an order $p$ tensor with rank $r$, the computational complexity of every iteration in \algMMT{} is $ \calO(rp/d)$ with $d$ as the discretization size. Hence,
the computational complexity of the fixed resolution training is
\eq{
T
&= \calO\brck{
\fr{1}{|\log \alpha|} \cdot \log\fr{1}{ (1-\alpha)\epsilon}
\cdot\brck{\fr{rp}{d}}
} \notag\\
&= \calO\brck{
\fr{1}{|\log \alpha|} \cdot \log  \fr{1 }{(1-\alpha)\epsilon}
\cdot\brck{\fr{rp}{(1-\alpha)^2\epsilon}}
}. \notag\qed
}
In the multi-resolution setting, the spatial weights that are learned can be seen as a distribution that is approximated by its values at finitely many points.
Given a spatial discretization $d$, we can construct an operator $F_d$ that learns discretized tensor weights. The next Lemma relates the estimation error with resolution:
\begin{lemma}\cite{nash2000multigrid}
For each resolution level $[d_0, d_1, \cdots, d_n]$, there exists a constant $C_1$ and $C_2$,	such that the fixed point iteration with the discretization size $d$ has an estimation error:
\eq{\label{lemma:disc}
F(\T{W}) - F_d(\T{W}) \leq C_1 + \alpha C_2 \|\T{W}\|_d.
}
\end{lemma}
%
\proof See \cite{nash2000multigrid} for the details. \qed

We have obtained the discretization error for the fixed point operation at any resolution. Next we analyze the number of iterations $t(d)$ needed at each resolution $d$ before fine-graining.
\begin{lemma}
\label{lemma:iter_level}
For every resolution $d \in [d_0, d_1, \cdots, d_m]$, there exists a constant $C'$, such that the number of iterations  $t(d)$ before fine-graining satisfies:
\eq{
t(d) \leq C'/\log |\alpha|.}
\end{lemma}
\proof We know that $d_0 = 2d_1=4d_2=\ldots$. At resolution $d$, by using the estimate from the last level as initialization, we have:
\eq{
\T{W}_{2d}^{t(2d)} = \T{W}_d^0.
}
where we use the subscript to index the solution at a given resolution. By combining Lemma \ref{lemma:disc} and the fine-graining criteria in (\ref{eqn:fine-grain}), we can guarantee the estimation error per iteration:
\eqn{
&\|F_d(\T{W}_d^0) - \T{W}_d^0\| = \|F_d(\T{W}_{2d}^{t(2d)} ) ) - \T{W}_{2d}^{t(2d)} \| \notag\\
&\leq \|F_d(\T{W}_{2d}^{t(2d)} ) -F(\T{W}_{2d}^{t(2d)} )\| + \|F(\T{W}_{2d}^{t(2d)} ) - \T{W}_{2d}^{t(2d)}  \|) \notag\\
&\leq (C_1 + \alpha C_2 \|\T{W}_{2d}^{t(2d)} \|)d \notag\\
& + \|F(\T{W}_{2d}^{t(2d)} ) -F_{2d}(\T{W}_{2d}^{t(2d)} )) \| + \|F_{2d}(\T{W}_{2d}^{t(2d)} ))- \T{W}_{2d}^{t(2d)} ) \| \notag\\
&\leq (C_1 + \alpha C_2 \|\T{W}_{2d}^{t(2d)} \|)d +(C_1 + \alpha C_2 \|\T{W}_{2d}^{t(2d)} \|)2d + \alpha\fr{C_0d}{\alpha (1-\alpha)}.
}
According to the fixed point iteration definition, we have:
\eq{
\|F_d(\T{W}^{t(d)}) - \T{W}^{t(d)}) \| \leq \alpha^{t(d)-1} \| F_d(\T{W}_d^0) - \T{W}_d^0  \| \leq C' d \nonumber
}
Thus the number of iterations satisfies $t(d) \leq C'/\log |\alpha|$ for all resolutions. \qed.
\paragraph{Proof of Theorem \ref{thm:mmt}.}
By combining Lemmas \ref{lemma:iter_level} and the computational cost per iteration, we can compute the total computational cost  for our \algMMT{} algorithm, which is proportional to the total number of iterations for all resolutions:
\eq{
T &= \calO\brck{\fr{1}{|\log \alpha|}\brcksq {(d_n/rp)^{-1} +(2 d_n/rp)^{-1} + (4 d_n/rp)^{-1} + \cdots} } \notag\\
&=\calO\brck{\fr{1 }{|\log \alpha| }\brck{\fr{rp}{d_n}}\brcksq{1 + \fr{1}{2} + \fr{1}{4} +\cdots} } \notag\\
&=\calO\brck{\fr{1 }{|\log \alpha|} \brck{ \fr{rp}{d_n} } \brcksq{ \fr{1-(\fr{1}{2}) ^{n}}{1-\fr{1}{2}} } } \notag\\
&= \calO\brck{\fr{1 }{|\log \alpha|} \brck{\fr{rp}{(1-\alpha)^2\epsilon}}},
}
where the last step uses the termination criterion in (\ref{eqn:fine-grain})
$d_n = \Omega ((1-\alpha)^2\epsilon )$. Comparing with the complexity analysis for the fixed resolution algorithm in Lemma \ref{lemma:fixed}, we complete the proof.
\qed
\begin{figure}[ht!]
\centering
\includegraphics[width=\linewidth]{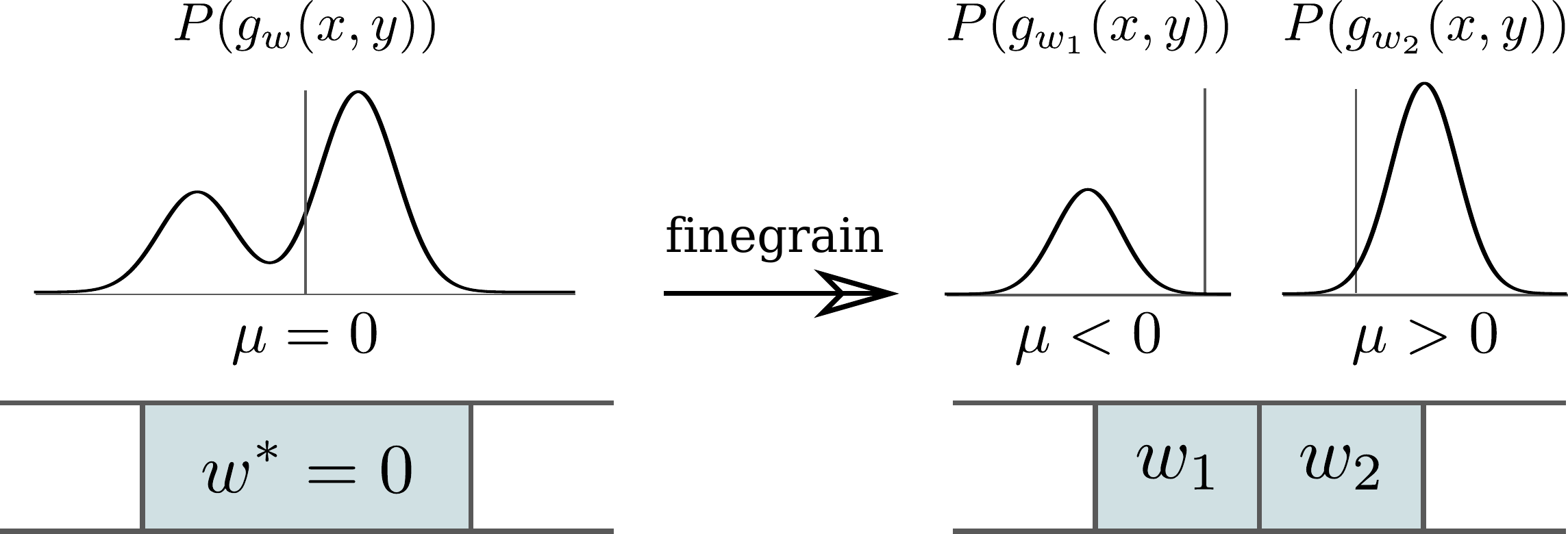}
\vspace{-0.05in}
\caption{Gradient distribution for a toy model during multi-resolution training. At the coarse level, the distribution has $\mu=0$ at convergence, whereas at a higher resolution, the distributions for the finegrained weights have non-zero mean.}
\label{fig:gradientdist}
\vspace{-0.13in}
\end{figure}
\subsection{Fine-Graining Criteria}
\label{ss:sgd}
Given the structure of \algMMT{} (Algorithm \ref{alg:mmt}), the key technical question is which fine-graining criterion to use at each stage. In Theorem \ref{thm:mmt}, we showed the improved convergence speed when using loss-convergence as a transition criterion. We now propose several fine-graining criteria to instantiate \algMMT{}, that empirically outperform loss-convergence.
\paragraph{Loss convergence.}
Intuitively, one simple fine-graining  condition is when training at the present resolution converges (which mimics the termination condition for iterative training of sparse models \cite{johnsonblitz}). That is, at training step $t$, we check whether:
\eq{\label{eq:loss_cond}
|\loss^t - \loss^{t}| < \tau_L,
}
where $\loss^{t}$ is the historical mean loss and $\tau_L$ is a threshold.
However, since the model at each resolution is used to initialize the training for the next resolution, the converged model at the current resolution might not be the best initialization for the next resolution (i.e., training might be overfitting to the coarser resolution).

\paragraph{Using Gradient Statistics.}
Given the spatial nature of the model weights \refn{eq:tucker}, we can use more finegrained information during training by analyzing the gradient distribution during training.
Intuitively, spatial resolution $d_i$ is too coarse when the data prefers much more fine-grained curvature, as evidenced by substantial disagreement in the gradients between resolution $d_i$ and $d_{i+1}$.
For example, consider learning a linear binary classifier $f$ as in \refn{eq:linear}:
\eq{
 \ww^* = \argmi{\ww} \mathE_{(x,y)\sim P}\brcksq{\loss(x,y,f(x;\ww))},
}
where $x\in\mathR$ and we consider two resolutions: $m_1 = 1$ cells and $m_{2} = 2$ cells. Suppose we optimize $\ww$ using 
%
%
%
\eq{
	\ww \leftarrow \ww + \Delta\ww, \hspace{5pt} \Delta \ww = - \lambda h(g_\ww; \theta),
}
where $\lambda$ is the learning-rate, $h$ is a (nonlinear) optimizer with parameters $\theta$ and the true gradient is
\eq{\label{eq:trueg}
 g_\ww = \
 \mathE_{(x,y)\sim P}\brcksq{ \nabla_\ww L(x,y,f(x;\ww))}.
}
Consider the situation that 1) during training at resolution $d_1$ the model has converged to $w^* = 0$ and 2) the ground truth model at resolution $d_{2}$ is $\ww^* = (+1,-1)$.
In this case, at the coarse resolution $d_1$ the distribution $P(g_\ww(x,y))$ of the point-wise gradient $g_\ww(x,y) = \nabla_\ww L(x,y,f(x;\ww))$ has mean $\mu\approx 0$ (and some variance $\sigma > 0$).
However, if we fine-grain the model, the (distribution of) gradients for the higher-resolutions weights $w_1$ and $w_2$ have non-zero mean: the mean gradient $g_{w_1}(x,y)$ will be negative and $g_{w_2}(x,y)$ positive. This is illustrated in Figure \ref{fig:gradientdist}.

More generally, we can quantify disagreement between gradients at multiple resolutions via the statistics $\mu, \sigma$ and entropy $S$ of the (empirical) gradient distribution of \ti{all} weights $\ww$ at resolutions $d_i$:
\eq{\label{eq:ent}
S\brck{g_{\ww\ddd{i}}} = \mathE_{ g_\ww(x,y) } \brcksq{ \log P\brck{g_\ww(x,y)}}.
}
Intuitively, when the entropy $S\brck{g_{\ww\ddd{i}}}$ is high, the gradients of the current discretization are increasingly disagreeing with each other and training can benefit from finegraining.
More formally, we can express this via the information gain:
\eq{\label{eq:ig}
 I_{i,i+1} = S\brck{g_{\ww\ddd{i}}} - S\brck{g_{\ww\ddd{i+1}}},
}
If $I_{i,i+1} > 0$, the gradient will have lower entropy at the higher resolution and finegraining could be beneficial.\footnote{This is analogous to the use of information gain for decision tree regularization.}

However, in practice it is infeasible to compute information gain, as this requires gradient statistics at a higher resolution.
As a proxy to \refn{eq:ig} we can instead use a simpler condition: finegrain when for $\geq p\%$ of the weights in $\ww$ the entropy $S$ exceeds a margin $\tau$:
\eq{\label{eq:ent_cond}
\textrm{for } \geq p \textrm{\% weights } w_j: S\brck{g_{w_j\ddd{i}}} > \tau,
}
where $\tau$ and $p$ are tunable hyperparameters.

\paragraph{Moment-based thresholds.}
We can define other criteria based on gradient statistics: finegrain if for at least $p\%$ of the weights:
\eq{\label{eq:mom_cond}
\sigma\textrm{-threshold: }& \sigma_{t} > \tau_\sigma, \\
\mu,\sigma\textrm{-threshold: }& \sigma_{t} > \tau_\sigma \textrm{ and } |\mu_{t}| < \tau_\mu.
}
where $\mu_{t}$, $\sigma^2_{t}$ are the gradient mean and variance at step $t$ and $\tau_\cdot$s are tunable hyperparameters.
$\sigma$-thresholding is a more coarse statistic compared to the entropy $S$: if the gradient distribution is non-Gaussian or multimodal $S$ can capture higher-order statistics as well. $\mu,\sigma$-thresholding also tracks whether training has converged ($\mu \approx 0$) according to the gradients.

\begin{table}[t]
  \centering
  \footnotesize
  \begin{tabular}{lc|lc}
    Name &   Range  & Name &  Range  \\
    \hline
    Learning-rate $\lambda$ & $10^{-5} - 10^{-1}$ &
    $\tau_l$ & $10^{-6}-10^{-2}$ \\
    Gradient buffer window $T$ & 10 - 1000 &
    $\tau_S$ & $10^{-6}-10^{-0}$ \\
    $L_1$-regularization  & $10^{-5} - 10^{-1}$ &
    $\tau_\mu$ & $10^{-8}-10^{-2}$ \\
    $L_2$-regularization  & $10^{-5} - 10^{-1}$ &
    $\tau_\sigma$ & $10^{-8}-10^{-1}$ \\
    Criterion check frequency  $\omega$ & $10 - 1000$ &&
  \end{tabular}
  \caption{Search range for \algSGD{} hyperparameters and fine-graining criterion hyperparamters.}
  \label{tab:hyperp}
\end{table}
\section{Benchmark Experiments}
\label{sec:quant}

To validate our approach, we demonstrate that our multi-resolution method converges faster than fixed-resolution training on two real-world datasets: basketball shots and fruit-fly behavior prediction.
Furthermore, we show that using spatial gradient statistics, such as its entropy, outperforms using loss convergence and provide a sensitivity analysis of the various transition criteria.
Finally, in Section \ref{sec:interpret}, we show that our learned models are interpretable.

For both datasets, we learn a tensor model for \ti{multi-task} prediction, where we instantiate the model \refn{eq:tucker} as a sum of low-rank $\T{W}^L$ and sparse $\T{W}^S$ tensors:
\eq{\label{eq:mt}
P( y_{a} | \xx ) &= \sum_{abc} (\T{W}^L_{abc} + \T{W}^S_{abc}) \phi_{b}(x) \psi_{c}(x),\\
\T{W}^L_{abc} &= \sum_{k=1}^K A_{ak}B_{bk}C_{ck}, \hspace{5pt} \T{W}^S_{abc} = \sum_{k=1}^K U_{ak}V_{bk}W_{ck},\notag
%
}
where $a\in\calA$ indexes the tasks.
A key motivation for this decomposition is that $L,S$ can capture semantically meaningful profiles (spatially dense and smooth, or sparse and peaky).

\begin{figure}[t!]
\centering
\includegraphics[width=0.15\textwidth]{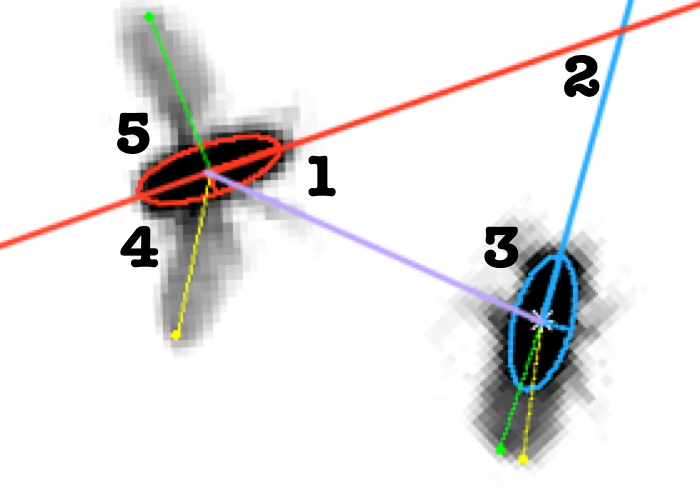}
\includegraphics[width=0.15\textwidth]{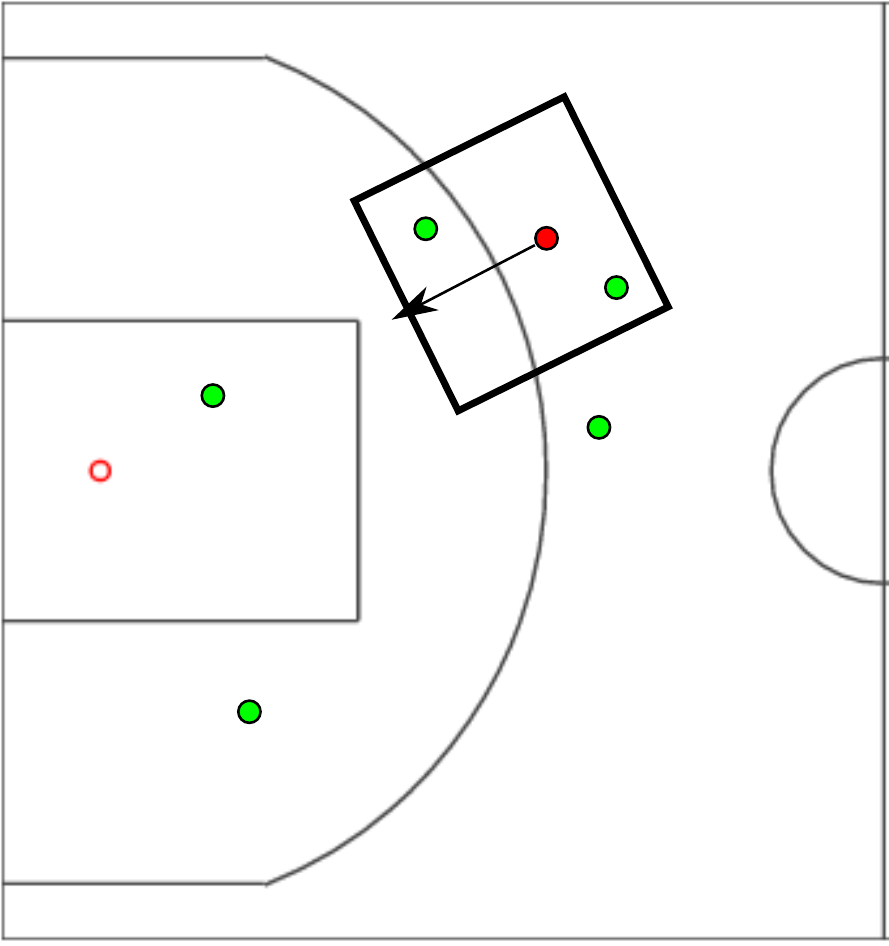}
\caption{Left: \FvF{} angles: \ti{facing angles} (1), (3), inter-fly \ti{angle} (2), \ti{minimal} (4) and \ti{maximal wing angle} (5)). Right: sample frame with ballhandler (red) and defensive players (green). Only players close to the ballhandler are used.}
\label{fig:fvf_features_explanation}
\end{figure}

\paragraph{Empirical Gradient Distribution.}
During training, we estimate the gradient distribution $P(g_\ww(x,y))$ by recording the empirical minibatch gradients and their statistics $\mu,\sigma,S$ over a rolling window of $T$ training steps. While collecting gradients, the weights $\ww$ are typically updated too, introducing a bias in the gradient statistics, as gradients are computed for different models at each step. However, we found empirically that using \refn{eq:ent_cond} remains effective.
\begin{figure*}[!h]
\centering
\includegraphics[height=115pt]{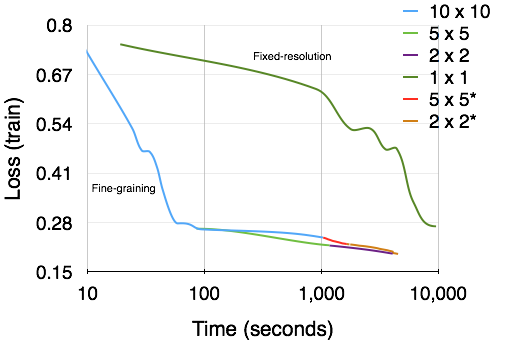}
\includegraphics[height=115pt]{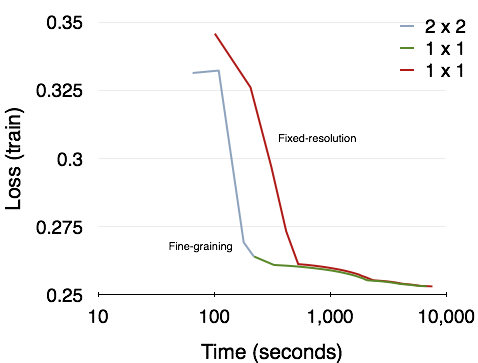}
\includegraphics[height=115pt]{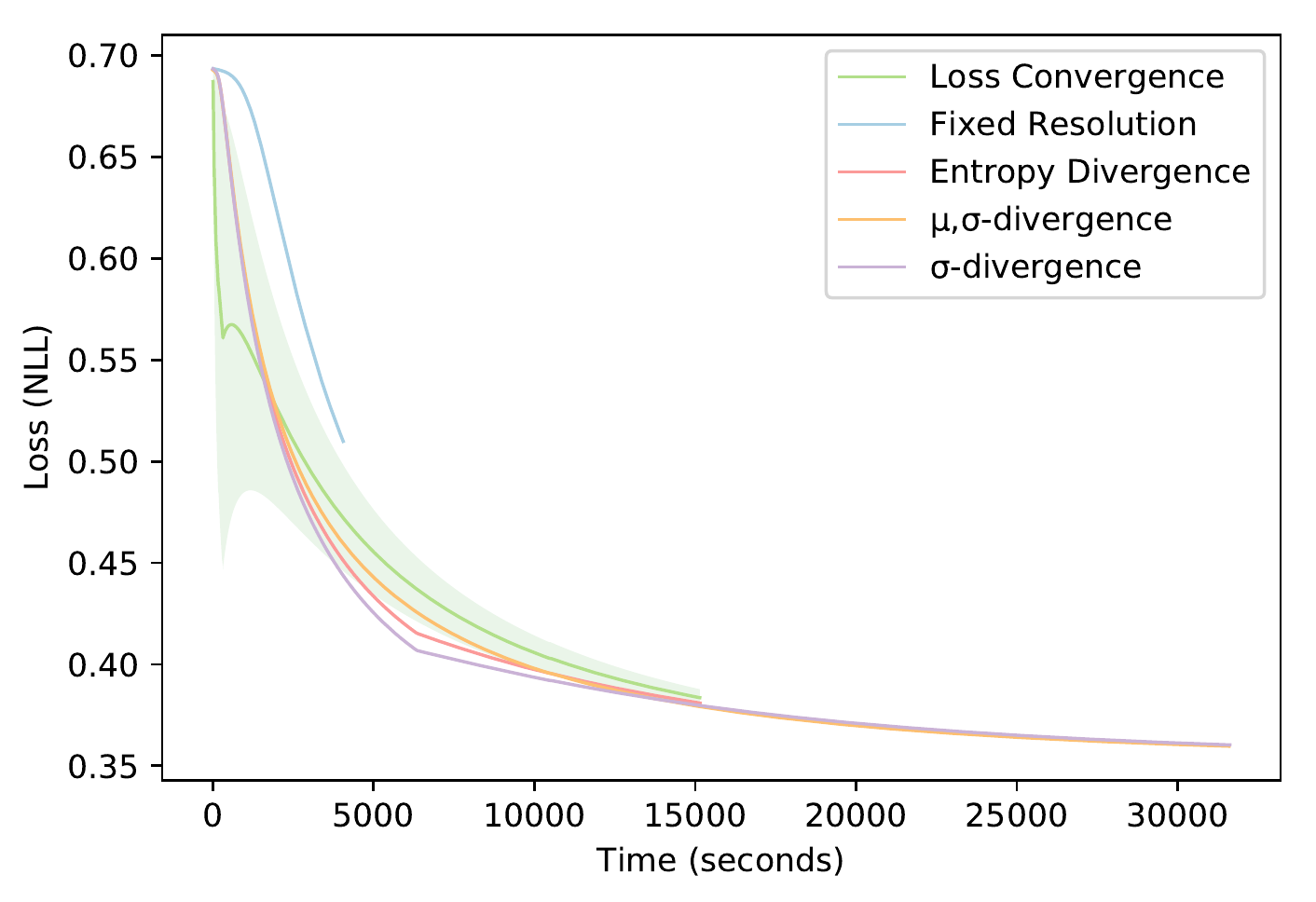}
\caption{Left: training a full-rank tensor model using \algMMT{} \refn{alg:mmt} with $\omega=100$ on \BB{}. Colors indicate training stages. For visual clarity, only best runs for fixed-resolution, loss-convergence and entropy-thresholding are shown.
Baseline 1 (dark green): training a fixed-resolution fine-grained model is orders of magnitude slower.
Baseline 2 (indicated by *): fine-graining using gradient entropy control (green, purple) with $20$ histogram bins converges faster than loss-convergence (red, orange).
Middle: training a factored tensor model using iteratively fine-graining outperforms a fixed-resolution approach.
Right: Learning a factorized model \refn{eq:mt} on \FvF{}. Using gradient statistics converges $\approx$ $50\%$ faster than fixed-resolution learning.
}
\label{fig:full_fixed_vs_cascade}
\end{figure*}

\paragraph{Experiments.}
First, we compared \algMMT{} with fixed-resolution training.
Secondly, we compared the various fine-graining criteria (entropy divergence, loss convergence, $\sigma$-threshold, $\mu,\sigma$-threshold) as described in section \ref{ss:sgd}.
We instantiated \algMMT{} by adaptively upscaling the model $B,C,V,W$ and features $\phi_b$ and $\psi_c$ as in Algorithm \ref{alg:mmt}, checking transition criteria each $\omega=100$ steps for $p=10\%$ and various $\tau$. All models were trained using Adam using cross-entropy loss and $L_2$ ($L_1$) regularization for the dense (sparse) factors.
We selected optimal hyperparameters for all models using a grid search over hyperparameters (see Table \ref{tab:hyperp}). For the sensitivity analysis, we used the $\tau$s as in Table \ref{tab:hyperp}.
%
%
%
%
\subsection{Datasets}
\paragraph{Basketball tracking data.}
We evaluated our method on \ti{basketball shot prediction}, where the tasks correspond to unique basketball players, indexed by $a$, and our multi-task model predicts \ti{whether player $a$ will shoot at the basket immediately after frame $\xx_\alpha$} ($\alpha$ indexes the dataset).
We used a large player tracking dataset \BB{} \cite{Yue2014,zheng2016generating} that includes hundreds of players and covers millions of game frames captured during competitive basketball gameplay. For each frame $\xx_\alpha$, we have binary labels $y_{\alpha,a} \in \{-1,+1\}$: whether player $a$ will (not) shoot at the basket in frame $\xx_\alpha$.
The features are 1-hot vectors $\phi(\xx_\alpha)\in\brckcur{0,1}^{2000}$ for the location $b$ of the ballhandler and $\psi(\xx_\alpha)\in\brckcur{0,1}^{144}$ for the defenders in a $12\times 12$ grid around the ballhandler. At full resolution, the court is discretized using a $50\times 40$ grid of 2000 cells of size $1\times 1$. For lower resolutions, we used $2\times 2, 3\times 3$ and $4\times 4$ cells.

\paragraph{Caltech Fly-vs-Fly.}
The Caltech Fly-vs-Fly behavior dataset \FvF{} \citep{Eyjolfsdottir2014} consists of several million video-frames of 2 fruit-flies, labeled by neuro-biologists with 10 behavioral fly actions: \ti{touch}, \ti{wing threat}, \ti{charge}, \ti{lunge}, \ti{hold}, \ti{tussle}, \ti{wing extension}, \ti{circle}, \ti{copulation attempt}, \ti{copulation}.
The goal is to predict whether either fly performs any of these 10 actions in video frame $\xx_\alpha$. Here, the tasks are the actions indexed by $a$ and the binary labels $y_{\alpha,a}\in\{-1,+1\}$ correspond to whether an action $a$ is present in frame $\xx_\alpha$ or not.
The dataset includes 12 pose and spatial features, including the \ti{velocity and the wing angles} of a single fly, and pairwise features, such as the \ti{distance and angle} between the two flies (see Figure \ref{fig:fvf_features_explanation}).
%
%
%
%
\begin{table}[t]
\centering
\resizebox{\linewidth}{!}{%
  \begin{tabular}{l|cc|cc}
    Loss-threshold: & 0.62 & Full-rank & 0.64 & Factor \\
    \hline
    Criterion & Minimal time (s) & Mean & Minimal & Mean \\
    \hline
    Fixed Resolution  & 18358 & - & 106 & - \\
    Loss-conv					& $639$ & $10141$ & $96$ & $99$ \\
    Entropy-threshold           & $541$ & $4258$ & $\bm{40}$ & $\bm{42}$ \\
    $\sigma$-threshold          & $356$ & $2453$ & $55$ & $56$ \\
    $\mu, \sigma$-threshold     & $\bm{292}$ & $\bm{677}$ & $880$ & $894$ \\
  \end{tabular}}
  \caption{Minimal and mean time to reach a loss when overfitting starts on \BBmini{} across 20 runs, with full and factorized model \refn{eq:tucker} (as in Figure \ref{fig:sensitivity}). Using gradient statistics reaches the loss threshold consistently significantly faster.}
  \label{tab:sens}
\end{table}
\subsection{Accuracy and Convergence Speed}
We compare \algMMT{} with fixed-resolution learning. Moreover, we compare a number of fine-graining criteria: fine-graining when the loss has converged versus spatial entropy control.
For the latter, we used condition \refn{eq:ent_cond} to determine when to fine-grain.
Figure \ref{fig:full_fixed_vs_cascade} and Table \refn{tab:sens} show our results using models with total latent dimension 20.\footnote{Performance trend is stable for a wide range of latent dimensions, omitted for brevity.}
The left plot shows the results for the full-rank model, which is usually the more computationally intensive part.
We see that our multi-stage approach dramatically outperforms (by multiple orders-of-magnitude) a naive approach which only uses the finest resolution.\footnote{Note that for more complex models, the naive approach would not even fit in memory.}
Moreover, spatial entropy control outperforms, by an order-of-magnitude, using the loss as a termination criterion.

The right plot shows the performance of \algMMT{} on the latent factor model after initializing using a factorized model of the previous stage. We see that the learning objective continues to decrease as learning enters what is essentially a fine-tuning phase.\footnote{Note the absolute objective of the un-factorized model is lower than the latent-factor model because the former has more degrees of freedom and is overfitting.}
\begin{figure}[t]
\centering
\includegraphics[width=0.68\linewidth]{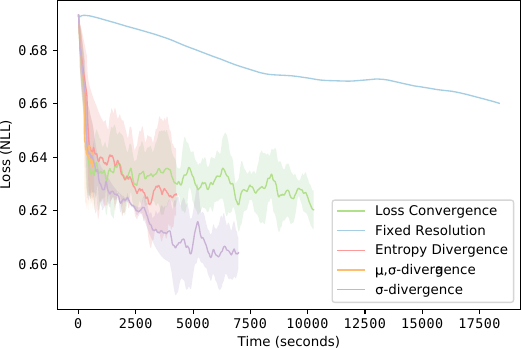} \\
\includegraphics[width=0.68\linewidth]{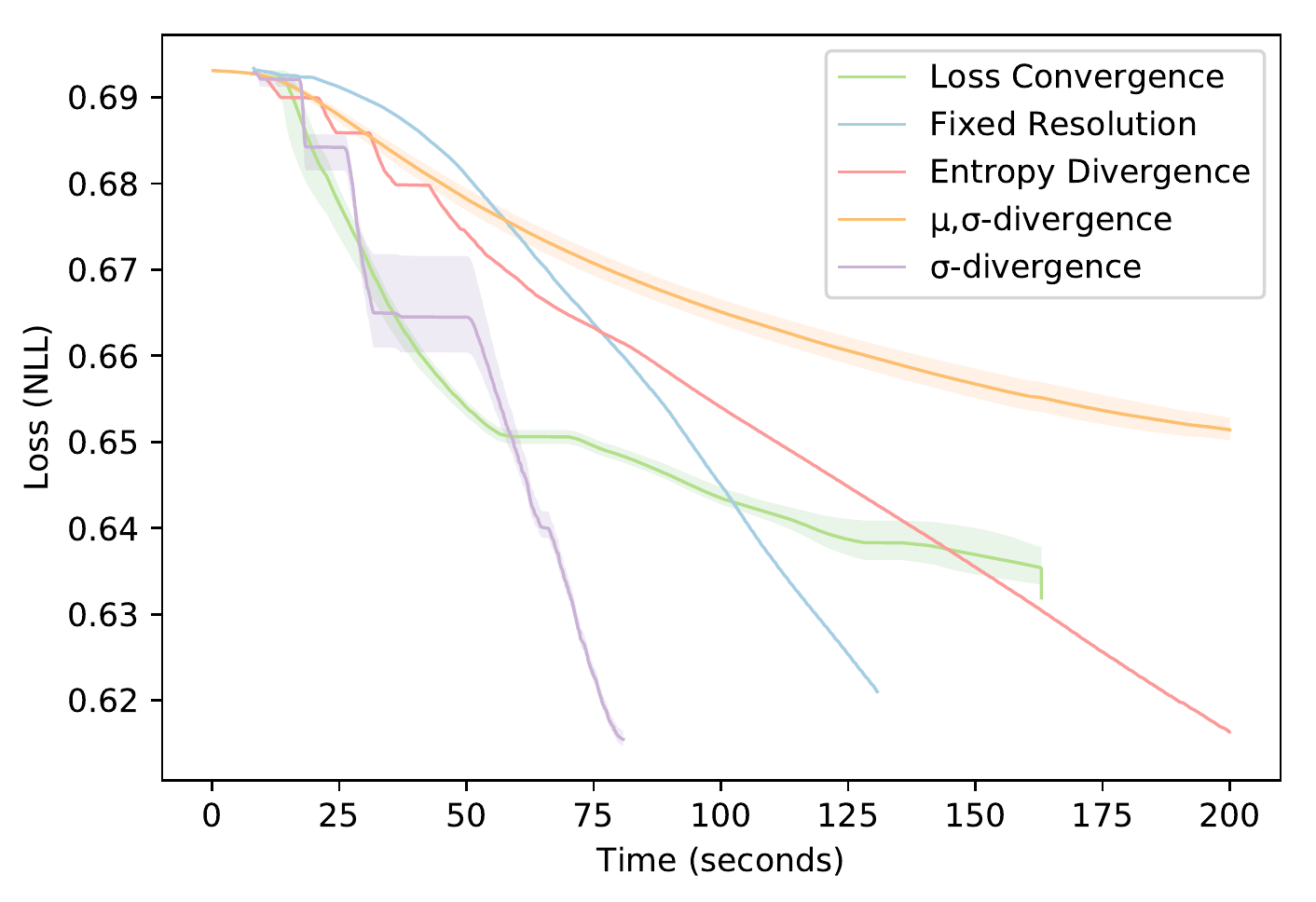}
\caption{Sensitivity of \algMMT{} for 4 transition criteria (e.g. \refn{eq:ent_cond}) to hyperparameters $\tau$ and $p=10\%$ on \BBmini{} using a full-rank (top) and a factored model \refn{eq:tucker} (bottom). For each criterion, we sampled 20 $\tau$s uniformly and show the mean and variance of the converging runs as they start to overfit.}
\label{fig:sensitivity}
\end{figure}
\subsection{Sensitivity Analysis}
To evaluate the efficiency and transition behavior using the transition criteria from Section \ref{ss:sgd}, we evaluated their sensitivity to the threshold parameters $\tau$. For this, we used \BBmini{} with 200k \BB{} examples and evaluated Algorithm \ref{alg:mmt} using a random search over $\tau$. The results are in Table \ref{tab:sens} and Figure \ref{fig:sensitivity}. We see that using gradient statistics (\refn{eq:ent_cond}, \refn{eq:mom_cond}) consistently outperforms loss convergence in the short-term ($\sigma$-threshold) and long-term (entropy-threshold). We empirically find that $\mu,\sigma$-threshold is harder to stabilize than other criteria, although it can outperform other methods.

\begin{figure}[t!]
\centering
\subfloat[ ]{%
\includegraphics[height=128pt]{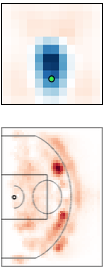}}
\label{fig:bb_lf_1}\hfill
\subfloat[ ]{%
\includegraphics[height=128pt]{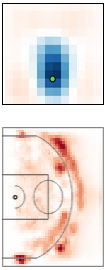}}
\label{fig:bb_lf_2}\hfill
\subfloat[ ]{%
\includegraphics[height=128pt]{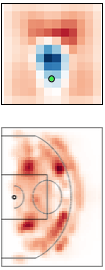}}
\label{fig:bb_lf_4}\hfill
\subfloat[ ]{
\includegraphics[height=128pt]{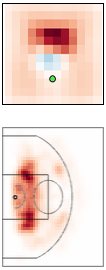}}
\label{fig:bb_lf_3}\\
\subfloat[ ]{%
\includegraphics[height=128pt]{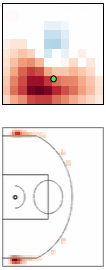}}
\label{fig:bb_lf_5}\hfill
\subfloat[ ]{%
\includegraphics[height=128pt]{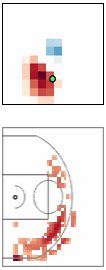}}
\label{fig:bb_lf_6}\hfill
\subfloat[ ]{%
\includegraphics[height=128pt]{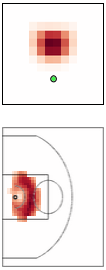}}
\label{fig:bb_lf_7}\hfill
\subfloat[ ]{%
\includegraphics[height=128pt]{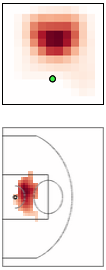}}
\label{fig:bb_lf_8}
\vspace{-0.1in}
\caption{
Row 1: smooth defender profiles $C_{ck}$ ($k = 1, 2, 3, 4$), row 3: sparse $W_{ck}$ ($k= 5, 6, 7, 8$) showing how defenders around the ballhandler influence the probability of a shot at the basket.
The ballhandler is located at $(6, 3)$ (green marker) and the vertical points towards the basket.
Row 2: smooth ballhandler shooting profiles $B_{bk}$ ($k = 1, 2, 3, 4$), row 4: sparse $V_{bk}$ ($k= 5, 6, 7, 8$).
The basket is located at the dot at (5.25, 20) and the 3-point line is the large arc.
All profiles are normalized (red: positive, blue: negative).
}
\label{fig:lfs_basketball}
\end{figure}
\section{Analysis of Interpretable Solutions}
\label{sec:interpret}
We now qualitatively evaluate a learned model \refn{eq:mt} that demonstrates that our meta-algorithm \algMMT{} (Algorihm \refn{alg:mmt}) can learn compact representations of semantic knowledge. We show that:
\iitem{
    \item The latent factors are interpretable: they capture characteristic basketball shooting profiles and spatial fly behavior.
    \item Smooth and sparse latent factors correspond to various types of basketball behavior and spatial configurations of fruit flies.
}
\subsection{Basketball shot prediction}
\paragraph{Ball handler shooting profiles.}
%
%
In competitive basketball play, shots at the basket happen throughout the court and peak at certain hot-spot locations at short range (close to the basket), medium range (between the basket and 3-point line) and long range (at the 3-point line or beyond).
Inspecting the smooth shooting profiles in Figure \ref{fig:lfs_basketball}, we see that all profiles are spatially cohesive and have small peaks around different subsets of hot-spots.
Profiles 1, 2 and 3 capture medium range and long range hot-spots, while profile 4 covers the short-range zone.
In contrast, the sparse shooting profiles are more spatially concentrated and activate only on specific hot-spots.
For example, profile 5 covers 2 hot-spots to the far left and right side at the back of the court and profiles 7 and 8 only activate on the short range zone directly around the basket.

\begin{figure}[h]
\centering
\includegraphics[width=0.8\linewidth]{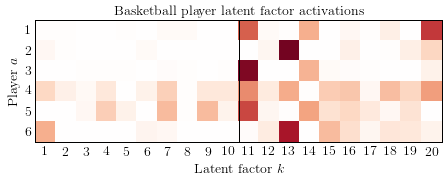}
\vspace{-10pt}
\caption{Basketball player latent factor activation weights $A_{ak}$ for the same model as in Figure \ref{fig:lfs_basketball} with 10 smooth factors ($k = 1\ldots 10$) and 10 sparse factors ($k = 11\ldots 20$).
}
\label{fig:basketball_activations}
\end{figure}

This leads to an attractive semantic interpretation: smooth shooting profiles describe \textit{inconsistent players} that tend to shoot from many locations throughout the basketball court, whereas sparse shooting profiles capture \textit{consistent players} that shoot from only specific hot-spots.
Figure \ref{fig:basketball_activations} show the latent factor activations $A_{ak}$ and $U_{a,k}$ of 6 players that can be grouped as such:
players 1, 2, and 3 are \ti{consistent}; players 4, 5, and 6 are more \ti{inconsistent}.


\paragraph{Defender influence profiles.}
%
%
The bottom row of Figure \ref{fig:lfs_basketball} depicts four dense defender profiles $C$ and four sparse defender profiles $W$. The first two dense profiles capture defender suppression in front of the ballhandler across the entire court, since the companion shooting profiles (1 and 2) are diffused throughout. In contrast, the first two sparse defender profiles (5 and 6) describe defender influence specifically at long range, which has a more peaked behavior.
The last two sparse profiles (7 and 8) describe ballhandlers that are prone to shoot with a defender close by. This is likely due to confusing correlation with causation, since players close to the basket tend to shoot, even though typically a defender is close by.
\subsection{Fruit fly behavior}
\begin{figure}[h]
\centering
\includegraphics[width=0.8\linewidth]{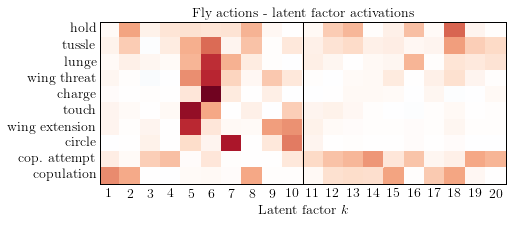}
\vspace{-10pt}
\caption{Fly action latent factor activation weights $A_{ak}$ for a model with 10 smooth ($k = 1\ldots 10$) and 10 sparse latent factors ($k = 11\ldots 20$). Similar actions activate similar factors.}
\label{fig:fvf_task_activations}
\end{figure}

\begin{figure}[t!]
\centering
\includegraphics[height=100pt]{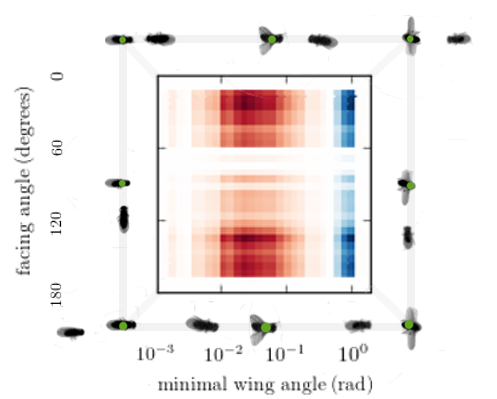} 
\includegraphics[height=100pt]{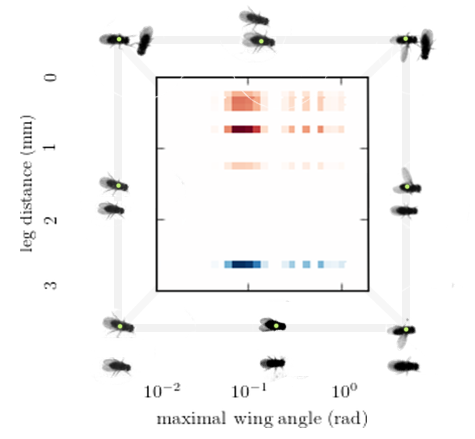}
\caption{
Left: Smooth normalized profile in $B_{bk}C_{ck}$ over \ti{facing angles} and \ti{wing angles}. Right: Sparse profile. Red (blue) means positive (negative).
Example spatial configurations of a pair of flies are shown for the extremal regions of the profile. The reference fly (green) always points to the right, with the second fly at various facing angles.
The dense profile describes a fly that extends exactly one wing, the maximal (minimal) wing length is large (small) and is turned towards the other fly.
The sparse profile activates highly when the legs of the two flies are close and the reference fly keeps both wings close to his body.
}
\label{fig:fvf_profile}
\end{figure}

\paragraph{Configuration profiles.}
Figure \ref{fig:fvf_profile} depicts a sample dense and a sparse learned profile that are interpretable. In the dense profile, the fly extends \ti{at least one wing} which defines \textit{wing extension}, a sign of courtship.
The sparse profile similarly captures more concentrated spatial characteristics, when the flies are close. Such wing configurations are important in both aggressive and courtship behavior, when the two flies interact closely to each other.
\paragraph{Actions and behaviorial types.}
Figure \ref{fig:fvf_task_activations} provides a more holistic view across all actions in the \FvF{} dataset. For example, we see that \ti{lunge} and \ti{wing threat} have similar activations. This is natural: the former is active physical aggression, while the latter can be interpreted as a fly showing a preview of physical aggression to intimidate the other fly.  We observe that both dense and sparse profiles are useful for modeling a wide range of actions/tasks.

\section{Discussion}

Our adaptive multi-resolution approach learns models at a fixed \ti{global} resolution at each stage. An open problem remains how to learn at different resolutions \ti{locally}, i.e. by finegraining individual cells rather than all. Such methods could be used to learn the dynamics in systems that intrinsically behave differently at different resolution, such as many dynamical systems. Local gradient statistics could be used in this case to decide when and which cells to finegrain to optimize model performance.


\bibliographystyle{ACM-Reference-Format}
\bibliography{refs,refs_vi,refs_rl}


\end{document}


\title{Multi-resolution Tensor Learning for Large-Scale Spatial Data }

\author{Stephan Zheng}
\affiliation{%
 \institution{California Institute of Technology}
 \streetaddress{1200 E California Boulevard}
 \city{Pasadena}
 \state{California}
 \postcode{91125}
}
\email{stephan@caltech.edu}
\author{Rose Yu}
\affiliation{%
 \institution{California Institute of Technology}
 \streetaddress{1200 E California Boulevard}
 \city{Pasadena}
 \state{California}
 \postcode{91125}
}
\email{rose@caltech.edu}
\author{Yisong Yue}
\affiliation{%
 \institution{California Institute of Technology}
 \streetaddress{1200 E California Boulevard}
 \city{Pasadena}
 \state{California}
 \postcode{91125}
}
\email{yyue@caltech.edu}

\renewcommand{\shortauthors}{S. Zheng et al.}

\maketitle

\section{Supplementary material}
\subsection{Computational Complexity Analysis}
\begin{lemma}
Given a fixed point iteration operator with a contraction factor of $\alpha \in (0,1)$, the computational complexity of a fixed-resolution training for a $P$ dimensional tensor with rank $r$ is
\[\calO\brck{
\fr{1}{|\log \alpha|} \cdot \fr{1 }{\log (1-\alpha)\epsilon}
\cdot\brck{\fr{1}{rp(1-\alpha)^2\epsilon}}}\]
\label{lemma:fixed}
\end{lemma}
\proof at high level, we prove this by choosing a small enough resolution $d$ such that the approximation error is bounded with a fixed number of iterations.

Let $\T{W}_d^\star$ be the optimal estimate at level $d$ and $\T{W}^t$ be the estimate at step $t$. Then we have
\eq{
\| \T{W}^\star -\T{W}^t \| \leq \| \T{W}^\star - \T{W}_d^\star \|  + \|\T{W}_d^\star - \T{W}^t  \| \leq \epsilon.
}
Choose a fixed resolution $d$ that is small enough such that
\eq{\| \T{W}^\star-\T{W}^\star_d\| \leq \fr{\epsilon}{2},}
then the termination criteria
\eq{
\| \T{W}^\star-\T{W}^\star_d\| \leq  \fr{C_0 d}{(1-\alpha)^2}
}
gives
\eq{
d = \Omega ((1-\alpha)^2 \epsilon).
}
Initialize $\T{W}^0 =0$ and iterate over $t$ times such that:
\eq{
\fr{\alpha^t}{2(1-\alpha) } \|T_d(\T{W}^0) \| \leq \fr{\epsilon}{2},
}
Given that $\T{W}^0 =0$, hence $\|T_d(\T{W}^0) \| \leq 2C$, we obtain that
\eq{
t\leq  \fr{1}{|\log \alpha|} \cdot \log\fr{2C}{(1-\alpha)\epsilon},
}
the computational complexity of the fixed resolution training is
\eq{
t
&= \calO\brck{
\fr{1}{|\log \alpha|} \cdot \fr{1}{\log (1-\alpha)\epsilon}
\cdot\brck{\fr{1}{drp}}
} \\
&= \calO\brck{
\fr{1}{|\log \alpha|} \cdot \fr{1 }{\log (1-\alpha)\epsilon}
\cdot\brck{\fr{1}{rp(1-\alpha)^2\epsilon}}
}.
}

\begin{lemma}\cite{nash2000multigrid}
For each resolution level $[d_0, d_1, \cdots, d_n]$, there exists a constant $C_1$ and $C_2$,	such that the fixed point iteration with discretization size $d$ has an estimation error:
\eq{\label{lemma:disc}
T(\T{W}) - T_d(\T{W}) \leq C_1 + \alpha C_2 \|\T{W}\|_d.
}
\end{lemma}
%
\proof The approximation error of a function with discretization is bounded if the target function is Lipschitz continuous. We have
\eq{
T(\T{W} + \Delta \T{W}) = T(\T{W}) +\alpha  \Delta \T{W},
}
and
\eq{
	\| T (\T{W} ) - T( \Delta \T{W}) ) \|\leq \alpha \| \T{W} -  \Delta \T{W} \|.
}
Here $T(\T{W}) $ is bounded by $C_1$ and $ \Delta \T{W}$ is bounded by $C_2 d$.
By definition of the fine-graining criterion, we know that for every resolution $d \in [d_0, d_1, \cdots, d_n]$, the discretized operator $T_d$ satisfies:
\eq{
\| T_d (\T{W} ) - \T{W}) \|\leq \fr{C_0d}{\alpha(1-\alpha)}.
}
%
%
If the estimation error is small enough, the algorithm terminates. Otherwise, we fine-grain and use $\T{W}^d$ to initialize the algorithm at the next resolution.
Let the termination criterion at level $d_n$ be
\eq{
\fr{C_0 d_n}{(1-\alpha)^2} \leq \fr{\epsilon}{2}.
}

\bibliographystyle{ACM-Reference-Format}
\bibliography{refs,refs_vi,refs_rl}
